# The Singularity May Be Near

**Roman V. Yampolskiy**

Computer Engineering and Computer Science
Speed School of Engineering
University of Louisville
roman.yampolskiy@louisville.edu

**Abstract**
Toby Walsh in "*The Singularity May Never Be Near*" gives six arguments to support his point of view that technological singularity may happen but that it is unlikely. In this paper, we provide analysis of each one of his arguments and arrive at similar conclusions, but with more weight given to the "likely to happen" probability.

## 1. Introduction

In February of 2016 Toby Walsh presented his paper "*The Singularity May Never Be Near*" at AAAI16 [23], which was archived on February 20, 2016 (http://arxiv.org/abs/1602.06462). In it, Walsh analyzes the concept of technological singularity. He does not argue that AI will fail to achieve super-human intelligence; rather he is suggesting that it may not lead to the runaway exponential growth. Walsh defends his view via six different arguments.

Almost exactly a year before, on February 23, 2015, Roman Yampolskiy archived his paper "*From Seed AI to Technological Singularity via Recursively Self-Improving Software*"[28] (http://arxiv.org/abs/1502.06512) which was subsequently published as two peer-reviewed papers at AGI15 [26, 29]. In it, Yampolskiy makes arguments similar to those made by Walsh, but also considers evidence in favor of intelligence explosion. Yampolskiy's conclusion is that Singularity may not happen but leans more toward it happening. In the next section, we present arguments from the original paper by Yampolskiy mapped to each of the six arguments given by Walsh in his work.

## 2. Contrasting Yampolskiy's and Walsh's Arguments

To make it easier to contrast arguments derived from *On the Limits of Recursively Self-Improving Artificially Intelligent Systems* [28] we use Walsh's naming of arguments even if our analysis doesn't rely on the same example (ex. No dog).

**Fast Thinking Dog**
Walsh argues: "… speed alone does not bring increased intelligence", and Yampolskiy says: "In practice performance of almost any system can be trivially improved by allocation of additional computational resources such as more memory, higher sensor resolution, faster processor or greater network bandwidth for access to information. This linear scaling doesn't fit the definition of recursive-improvement as the system doesn't become better at improving itself. To fit the definition the system would have to engineer a faster type of memory not just purchase more memory units of the type it already has access to. In general hardware improvements are likely to speed up the system, while software improvements (novel algorithms) are necessary for achievement of meta-improvements." It is clear from the original paper that performance in this context is the same as intelligence and as most of our intelligence testing tools (IQ tests) are time based, increased speed would in fact lead to higher Intelligence Quotient, at least in terms of how we currently access intelligence.

**Anthropocentric**
Walsh argues: "… that human intelligence is itself nothing special", and Yampolskiy says: "We still don't know the minimum intelligence necessary for commencing the RSI [Recursive Self-Improvement] process, but we can [argue] that it would be on par with human intelligence which we associate with universal or general intelligence [13], though in principal a sub-human level system capable of self-improvement can't be excluded [6]. One may argue that even human level capability is not enough because we already have programmers (people or their intellectual equivalence formalized as functions [18] or Human Oracles [24, 25]) who have access to their own source code (DNA), but who fail to understand how DNA (nature) works to create their intelligence. This doesn't even include additional complexity in trying to improve on existing DNA code or complicating factors presented by the impact of learning environment (nurture) on development of human intelligence. Worse yet, it is not obvious how much above human ability an AI needs to be to begin overcoming the "complexity barrier" associated with self-understanding."

**Meta-intelligence**
Walsh argues: "…strongest arguments against the idea of a technological singularity in my view is that it confuses intelligence to do a task with the capability to improve your intelligence to do a task" and cites a quote from Chalmers [6] as an example - "If we produce an AI by machine learning, it is likely that soon after we will be able to improve the learning algorithm and extend the learning process, leading to AI+". Yampolskiy says: "Chalmers [6] uses logic and mathematical induction to show that if an $AI_0$ system is capable of producing only slightly more capable $AI_1$ system generalization of that process leads to superintelligent performance in $AI_n$ after n generations. He articulates, that his proof assumes that the *proportionality thesis,* which states that increases in intelligence lead to proportionate increases in the capacity to design future generations of AIs, is true."

**Diminishing returns**
Walsh argues: "There is often lots of low hanging fruit at the start, but we then run into great difficulties to improve after this. … An AI system may be able to improve itself an infinite number of times, but the extent to which its intelligence changes overall could be bounded." Yampolskiy says, "… the law of diminishing returns quickly sets in and after an initial significant improvement phase, characterized by discovery of "low-hanging fruit", future improvements are likely to be less frequent and less significant, producing a Bell curve of valuable changes."

**Limits of intelligence**
Walsh argues: "There are many fundamental limits within the universe", Yampolskiy outlines such limits in great detail "First of all, any implemented software system relies on hardware for memory, communication and information processing needs even if we assume that it will take a non-Von Neumann (quantum) architecture to run such software. This creates strict theoretical limits to computation, which despite hardware advances predicted by Moore's law will not be overcome by any future hardware paradigm. Bremermann [4], Bekenstein [2], Lloyd [12], Anders [16], Aaronson [1], Shannon [19], Krauss [10], and many others have investigated ultimate limits to computation in terms of speed, communication and energy consumption with respect to such factors as speed of light, quantum noise, and gravitational constant." "In addition to limitations endemic to hardware, software-related limitations may present even bigger obstacles for RSI systems. Intelligence is not measured as a standalone value but with respect to the problems it allows to solve. For many problems such as playing checkers [17] it is possible to completely solve the problem (provide an optimal solution after considering all possible options) after which no additional performance improvement would be possible [14]."

**Computational complexity**
Walsh argues: "… no amount of growth in performance will make undecidable problems decidable" and Yampolskiy says, "Other problems are known to be unsolvable regardless of level of intelligence applied to them [22]. Assuming separation of complexity classes (such as P vs NP) holds [27], it becomes obvious that certain classes of problems will always remain only approximately solvable and any improvements in solutions will come from additional hardware resources not higher intelligence."

## 3. Response to Walsh's Arguments
In this section we provide novel analysis of all six arguments presented by Walsh and, via mapping provided in the previous section, revisit and critically analyze arguments made by Yampolskiy.

**Fast Thinking Dog**
The argument intuitively makes sense, since nobody ever managed to train a dog to play chess. However, intuition is no match for a scientific experiment. Animals have successfully been trained to understand and even use human (sign) language and do some basic math. People with mental and learning disabilities, who have been long considered a "lost cause", have been successfully trained to perform very complex behaviors via alternative teaching methods and longer training spans. It is entirely possible that if one had thousands of years to train a dog it would learn to play a decent game of chess, after all it has a neural network very similar to the one used by humans and deep learning AI. It may be argued, that there is considerable evidence that language and some other capabilities are functions of specific brain structures that are largely absent from a dog. Thus, 1000s of years of training won't cut it and one would need millions of years of evolution to get a human-level intelligent dog. However some recent research has documented that people missing most of their brain could have near normal cognitive capacity [8] and even significant damage to parts of the brain could be overcome due to neuroplasticity [9] suggesting that brain structures are much more general. To transfer an analogy to another domain, Intel286 processor is not fast enough to do life speech recognition, but if you speed it up it is. Until an actual experiment can be performed on an accurately simulated digital dog, this argument will remain as nothing but speculation.

## Anthropocentric

The reason some experts believe ([3] - page 339; [11] - chapter 3) that human level of intelligence is special is not because of anthropocentric bias, but because of the Church-Turing Thesis (CTT). CTT states that a function over natural numbers is computable by a prototypical human being if and only if it is computable by some Turing Machine (TM), assuming such theoretical human has infinite computational resources similar to an infinite tape available to a TM. This creates equivalence between human level intelligence and a Universal Turing Machine, which is a very special machine in terms of its capabilities. However, it is important to note that the debate regarding provability of the CTT remains open [20, 5].

## Meta-intelligence

If the system is superior to human performance in all domains, as required by definition of superintelligence, it would also be superior in the domain of engineering/computer science/AI research. Potentially, it would be capable of improving intelligence of its successor up to any theoretical/physical limits which might represent an upper bound on optimization power. In other words, if it is possible to improve intelligence a superintelligent system will do so, but as such possibility remains a speculation, this is probably the strongest of all presented objections to intelligence explosion.

## Diminishing returns

It is a mathematical fact that many functions, while providing diminishing returns, continue diverging. For example, harmonic series: $1 + 1/2 + 1/3 + 1/4 + 1/5 + \ldots = \infty$, which is a highly counterintuitive result, yet is a proven mathematical fact. Additionally, as the system itself would be improving it is possible that the discoveries it will make with respect to future improvements will also improve in terms of their impact on the overall intelligence of the system. So while it is possible that diminishing returns will be encountered it is just as possible that returns will not be diminished.

## Limits of intelligence

While physical and theoretical limits to intelligence definitely exist they may be far beyond our capacity to get to them in practice and so will have no impact on our perception of machine intelligence appearing to be undergoing intelligence explosion. It is also possible that physical constants are not permanently set, but dynamically changing which has been demonstrated for some such physical "constants". It is also possible that the speed of improvement in intelligence will be below the speed with which some such constants will change. To bring an example from another domain, our universe can be said to be expending faster than the speed of light, with respect to distance between some selected regions, so even with travel at maximum theoretical speed (of light), we will never hit a limit/edge. So, again this is another open questions and limit may or may not be encountered in the process of self-improvement.

## Computational complexity

While it is certainly true that undecidable problems will remain undecidable, it is not a limitation on intelligence explosion as not a requirement to qualify as superintelligent and plenty of solvable problems exists at all levels of difficulty. Walsh correctly points out that most limitations associated with computational complexity are only problems with our current models of computation and are avoided by switching to different paradigms of computation such as quantum computing and some, perhaps not yet discovered, implementations of hupercomputation.

## 4. Conclusions

Careful side-by-side analysis of papers by Walsh and Yampolskiy shows an almost identical set of arguments against possibility of technological singularity. This level of successful replication in analysis is an encouraging fact in science and gives additional weight to shared conclusions, but in this paper we provide novel analysis of Walsh's/Yampolskiy's arguments which shows that they may not be as strong as initially appears. Future productive directions of analysis may concentrate on a number of inherent advantages, which may permit AI to recursively self-improve [21] and possibly succeed in this challenging domain: ability to work uninterrupted (no breaks, sleep, vocation, etc.), omniscience (complete and cross disciplinary knowledge), greater speed and precision (brain vs processor, human memory vs computer memory), intersystem communication speed (chemical vs electrical), duplicability (intelligent software can be copied), editability (source code unlike DNA can be quickly modified), near-optimal rationality (if not relying on heuristics) [15], advanced communication (ability to share cognitive representations complex concepts), new cognitive modalities (sensors for source code), ability to analyze low level hardware, ex. individual registers, addition of hardware (ability to add new memory, processors, etc.) [30]. The debate regarding possibility of technological singularity will continue. Interested readers are advised to read the full paper by Yampolskiy [28] as well as a number of excellent relevant chapters in Singularity Hypothesis [7] which address many arguments not considered in this paper.

## Acknowledgements

Author wishes to thank Toby Walsh for encouraging and supporting work on this paper as well as reviewers who provided feedback on an early draft and by doing so made the arguments presented in the paper much stronger.